# A Cost-Effective Framework for Predicting Parking Availability Using Geospatial Data and Machine Learning

Madyan Bagosher [1], Tala Mustafa [1], Mohammad Alsmirat [2,3], Amal Al-Ali [1], Isam Mashhour Al Jawarneh [1]

[1] Department of Computer Science, University of Sharjah, P.O.Box. 27272 Sharjah, United Arab Emirates
aialali@sharjah.ac.ae, ijawarneh@sharjah.ac.ae

[2] Department of Computer Science and Information Systems, East Texas A&M University, Commerce, Texas, USA
Mohamamd.Alsmirat@tamuc.edu

[3] Department of Computer Science, Jordan University of Science and Technology, Irbid P. O. Box 27272, Jordan
masmirat@just.edu.jo

*Abstract*— As urban populations continue to grow, cities face numerous challenges in managing parking and determining occupancy. This issue is particularly pronounced in university campuses, where students need to find vacant parking spots quickly and conveniently during class timings. The limited availability of parking spaces on campuses underscores the necessity of implementing efficient systems to allocate vacant parking spots effectively. We propose a smart framework that integrates multiple data sources, including street maps, mobility, and meteorological data, through a spatial join operation to capture parking behavior and vehicle movement patterns over the span of 3 consecutive days with an hourly duration between 7AM till 3PM. The system will not require any sensing tools to be installed in the street or in the parking area to provide its services since all the data needed will be collected using location services. The framework will use the expected parking entrance and time to specify a suitable parking area. Several forecasting models, namely, Linear Regression, Support Vector Regression (SVR), Random Forest Regression (RFR), and Long Short-Term Memory (LSTM), are evaluated. Hyperparameter tuning was employed using grid search, and model performance is assessed using Root Mean Squared Error (RMSE), Mean Absolute Error (MAE) and Coefficient of Determination ($R^2$). Random Forest Regression achieved the lowest RMSE of 0.142 and highest $R^2$ of 0.582. However, given the time-series nature of the task, an LSTM model may perform better with additional data and longer timesteps.

## I. Introduction

A critical challenge in today's smart cities is that of managing parking occupancy, attributed mostly to rapid urbanization. The inability of cities in fulfilling driver's needs and to efficiently allocate vacant parking slots potentially leads to blocked roads, traffic congestion, and delays for drivers. It also worsens broader urban issues such as noise pollution, improper parking, and increased accident risks. Studies have shown that U.S. drivers spend an average more than 15 hours annually searching for parking, while drivers in other countries (such as the U.K. and Germany) spend even greater amount of time (more than 40 hours per year). This directly contributes to significant economic burdens, with drivers wasting a lot of money yearly searching for parking spaces, aggregating to a staggering €40 billion annual cost to some national economies. These figures rationale the pressing need for novel solutions to address parking management effectively.

The hybridization of Internet of Things (IoT) and Web of Things (WoT) technologies has revolutionized urban systems employed in dynamic smart city applications, by basically enabling real-time data collection and analysis through connected sensors and devices. IoT-enabled sensors, with machine learning models, are considered promising for addressing urban issues such as traffic management and the optimization of parking slots allocation. However, the sole dependence on costly sensor infrastructure, such as in-situ sensors or surveillance cameras installed on streets and parking areas, can potentially limit the scalability of the system, and often can raise privacy issues. Having said that, there is a growing need for novel approaches that leverage publicly available geospatial data and machine learning predictive analytics to serve cost-effective and privacy-preserving parking slot's allocation solutions.

In this paper, we propose a novel framework that integrates several georeferenced data sources, including street maps, mobility patterns, and parking data, using spatial join operations to analyze parking behavior and vehicle movement dynamics. Unlike conventional systems that otherwise rely on costly hardware installations, our approach eliminates the need for specialized sensing equipment, depending instead on aggregated location-based data from public platforms such as OpenStreetMap (OSM), Google Maps, and other public location-based services. By predicting optimal parking locations based solely on the user's parking entry point and arrival time, the framework offers a scalable and sustainable solution for smart parking management.

To validate the effectiveness of the proposed framework, we tested the framework with several machine learning models, including Linear Regression, Support Vector Regression (SVR), Random Forest Regression (RFR), and Long Short-Term Memory (LSTM). Hyperparameter tuning using grid search was applied, and the performance was evaluated using metrics such as Root Mean Squared Error (RMSE), Mean Absolute Error (MAE) and $R^2$. Our results demonstrate the feasibility of achieving high prediction accuracy without the need for costly infrastructure and sensor or camera installations, ensuring the potential of geospatial data and machine learning to design novel urban parking systems.

The importance of our work presented in this paper extends beyond parking spaces management, because it aligns with a broader vision of smart cities aiming to improve urban living by exploiting and designing smart and sustainable technological systems for the management of cities. By addressing the challenges of parking availability in a cost-effective and privacy-preserving way, this novel framework potentially contributes to reducing traffic congestion, improving citizen satisfaction, and supporting smarter urban planning.

The rest of the paper is organized as follows: Section II covers the related works in the field; Section III defines the methodology; Section IV details the results and their discussion; and Section V concludes with the findings of this paper and future work perspectives.

## II. RELATED WORKS

Several approaches exist when determining parking occupancy, an Internet of Things (IoT) based solution proposed by Mohammed Balfaqih et al. integrates features such as: location tracking, parking management, real-time invoicing, and Wi-Fi transmission [1]. Similarly, Bala Murugan et al. used IoT devices such as: low cost sensors and Arduino microcontrollers to detect parking vacancy and performed predictive analysis in RStudio [2]. However, these solutions introduce privacy and security concerns that need to be addressed [3].

A different approach involves utilizing unstructured data such as: images and videos paired with powerful CNN models for detecting parking occupancy. Jasvina Davan et al. uses an object detection model called You Only Look Once YOLOv3 [4] and in-house data to detect parking occupancy in First City University College (FCUC) in Malaysia [5]. A similar study uses YOLOv4 [6] to identify parking occupancy and Support Vector Regression (SVR) to predict the number of vacant parking spaces available [7]. While these models are accurate, they do have several weaknesses [8] and they also require more training data and computational resources during training.

Several other studies were also conducted within university campuses. Akram Elomiya et al. proposed a smart parking system that uses Mamdani type Fuzzy Inference System (FIS). The experimental validation of the model was conducted at Pardubice University's campus in the Czech Republic and the model displayed an accuracy of 92% [9]. A recent study by Sohil Paudel et al. at the University of Texas at Tyler used Linear Regression and two artificial neural networks (ANNs) to address campus parking demand. One ANN, with an R-squared value of 0.846, provided an equation that course schedulers could use to optimize schedules and reduce parking demand [10].

The main issue with many of these studies is the collection of sensitive data, such as vehicle models and make, license plate numbers, and arrival/departure times from the campus. Also, the need for special sensing equipment that need to be installed in the parking areas which will incur technical difficulties and additional cost. Additionally, many of the deep learning models used require vast amounts of data and significant computational resources to achieve high accuracy. In this study, we propose a more efficient approach that focuses solely on monitoring the flow of vehicles in and out of campus gates using location services, leveraging machine learning models to predict parking occupancy. Furthermore, an interactive website has been developed to assist students in locating available parking spots on the university campus. This tool enhances campus navigation by providing real-time parking availability, reducing the time spent searching for vacant spaces.

## III. PROBLEM FORMULATION AND METHODOLOGY

The proposed framework offers a privacy-preserving and cost-efficient approach to predicting parking availability, eliminating the need for expensive camera or sensor installations and the associated concerns of recording sensitive personal information. Instead, the system relies on tracking the influx and outflux of vehicles to make accurate predictions. This approach ensures the privacy of individuals within the university while significantly reducing infrastructure and maintenance costs. The architecture of the system begins with the integration of multiple data sources, including street data, mobility data, and parking data. These datasets are unified through a spatial join operation, creating a comprehensive dataset that captures the dynamics of parking behavior and vehicle movement patterns. The consolidated dataset is then utilized to train several predictive models, ensuring a robust evaluation process to identify the most accurate and reliable model for predicting parking availability. The proposed architecture in Fig. 1 depicts a high-level abstraction of the system pipeline.

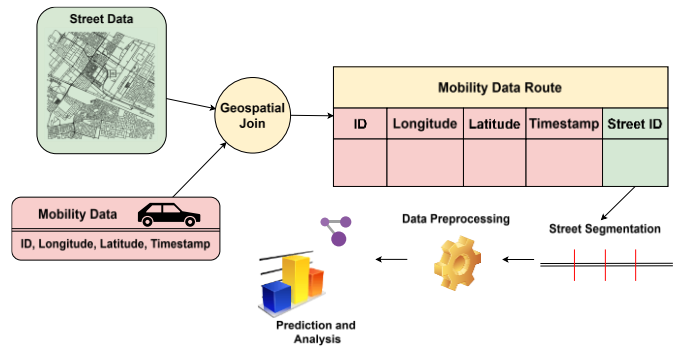

Fig. 1. Proposed Architecture of System

### A. Data Collection

Accurate data acquisition is essential for determining the availability of parking spaces. To analyze vehicular movement within the UOS campus, data delineating the campus road, surrounding road network and parking spots was obtained as a GeoDataFrame using OpenStreetMap (OSM). OSM is a free, collaborative, and editable geographic database of the world, created and maintained by a global community of volunteers. Often called the "Wikipedia of maps," OSM provides open-access spatial data, including roads, buildings, landmarks, and points of interest, under the Open Database License (ODbL). Unlike proprietary mapping services, OSM allows users to contribute, modify, and distribute its data for any purpose, making it a valuable resource for navigation, urban planning, disaster response, and research. The project relies on crowdsourced inputs from GPS traces, satellite imagery, and local knowledge, ensuring frequent updates and high accuracy in well-mapped regions.

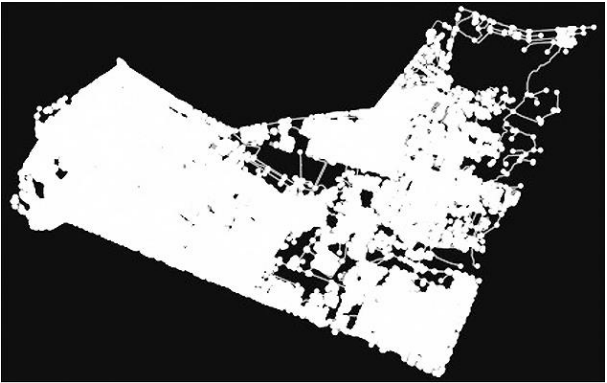

Fig. 2. UOS Road network visualized using OSMnx

The Python library OSMnx [11] was utilized to visualize the campus road network such as in Fig. 2, enabling comprehensive analysis. Additionally, mobility data was collected through web scraping using Outscraper, a powerful tool designed to extract real-time data. We utilized outscraper to obtain the geographical coordinates and timestamp associated with each vehicle within the campus road network by specifying the start and end points, time frame and time interval. Furthermore, information regarding the number of parking spots in each area of the campus was gathered and recorded manually. To enrich the dataset, a geospatial join operation [12] was performed to combine the two geo-referenced data sources. This process combined vehicular mobility data, geographic information about the UOS road network, and parking data. The resulting dataset provides a robust foundation for analyzing vehicular movement near the university campus and managing parking occupancy effectively. The dataset is comprised of 180 observations collected over the span of 3 consecutive days between 2022-09-05, 07:00:00 till 2022-09-08, 15:00:00 at an hourly rate since undergraduate courses are usually offered between these timings. Furthermore, the dataset is characterized by 25 distinct attributes and following an initial feature selection process, 17 attributes such as the road distance, query origin, query destination etc. were deemed redundant for the predictive task and were subsequently discarded. The remaining 8 attributes which are: Distance, Timestamp, Travel Speed, No. of Vehicles, No. of Vehicles exit, No. Segment, Total Parking Space and Availability were identified as relevant and informative and therefore retained for training the predictive models. Furthermore, the parking capacity varies across 3 campus sections, with the total parking capacity amounting to 945. To further enhance the quality and reliability of the dataset, additional preprocessing steps were undertaken. These steps included handling missing or inconsistent data, removing duplicates, one hot encoding categorical data and standardizing attribute formats where necessary. As a result of these preprocessing efforts, the dataset was refined to 144 observations, ensuring that only clean, high-quality, and representative data were used for model training. Finally, the streets were segmented between each of the campus gates to monitor the vehicle influx and outflux at each of the 5 gates. The segmentation process was critical in determining parking occupancy at each section across the campus as illustrated in Fig. 3. In the illustration Car 1 is in the 1$^{st}$ segment and is expected to enter Gate 1 unless it decides to proceed to segment 2 and Car 2 is expected to enter Gate 2 while it is positioned within the 2$^{nd}$ segment.

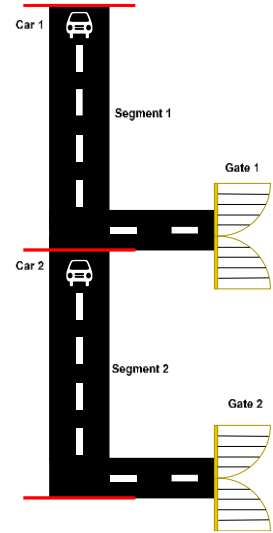

Fig. 3. Road Segmentation

### B. Data Analysis and Feature Selection

To ensure that the influx and outflux data accurately represent parking availability, a detailed statistical analysis was conducted. The Pearson correlation coefficient ($r$) was first calculated to evaluate the linear relationship between outflux and parking availability, yielding a value of 0.41 as shown in Fig. 4.

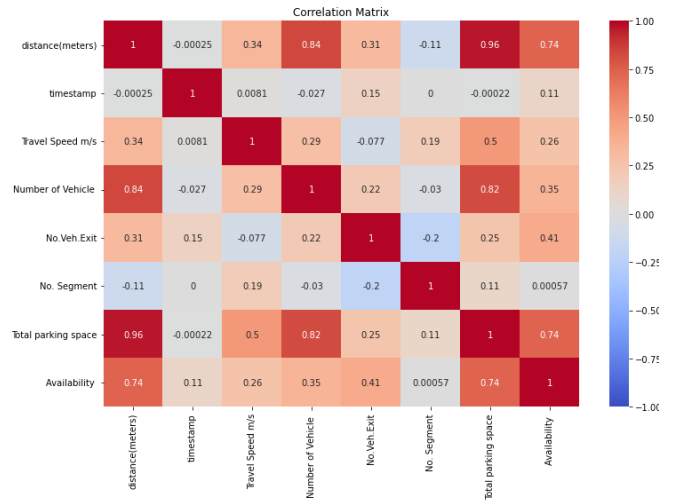

Fig. 4. Pearson Correlation coefficients

$$r = \frac{\sum(x_i - \bar{x})(y_i - \bar{y})}{\sqrt{\sum(x_i - \bar{x})^2 \sum(y_i - \bar{y})^2}} \quad (1)$$

This result indicates a moderate positive correlation, suggesting that an increase in outflux is moderately associated with changes in parking availability. Subsequently, the Spearman rank correlation coefficient was computed to assess the monotonic relationship between these variables. The Spearman rank ($\rho$) was found to be 0.54, accompanied by a p-value of $6.63 \times 10^{-15}$, indicating a statistically significant and moderately strong positive correlation. This further supports the hypothesis that the outflux impacts parking availability.

$$\rho = 1 - \frac{6 \sum d^2}{n(n^2 - 1)} \quad (2)$$

To validate these findings, the features were visualized through appropriate graphical methods, confirming the influence of outflux on parking availability as shown in Fig. 4. These analyses collectively demonstrate that outflux is a meaningful predictor of parking availability, reinforcing its inclusion in the predictive modeling framework and Fig. 6 illustrates how the distribution of various features reveals important patterns in the frequency of their values.

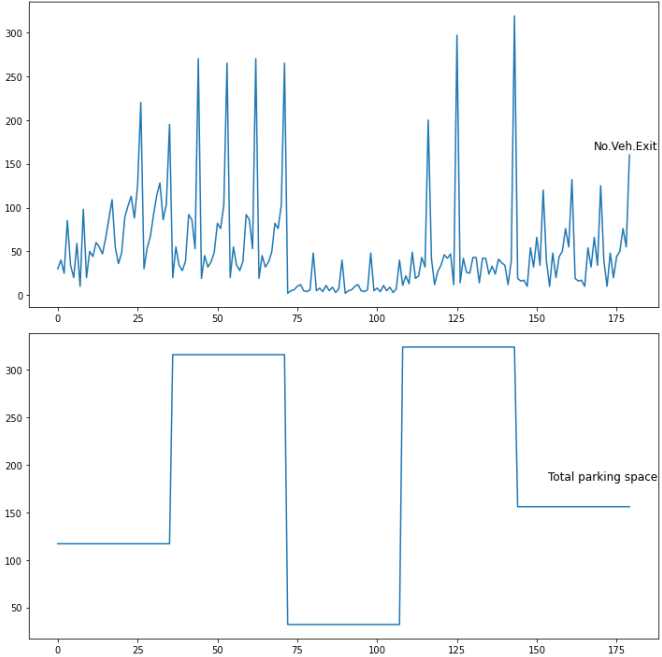

Fig. 5. Outflux and Parking Availability Time Series analysis

## C. Training Prediction Models

Four regression models were implemented and evaluated for predictive performance: Linear Regression, Long Short-Term Memory (LSTM) [13], Support Vector Regression (SVR) [14], and Random Forest Regression (RFR) [15]. The dataset was divided into a 70/30 train-test split to ensure robust model training and evaluation. Hyperparameter tuning was conducted using techniques like Grid Search and Random Search to identify the optimal parameter configurations for each model. Furthermore, to enhance generalization and prevent overfitting, 3-fold cross-validation was applied at every step and regularization techniques such as L1 and L2 were incorporated where necessary to improve model's predictions. The LSTM training graph in Fig. 8 displays the MSE of the training and testing data across 50 epochs.

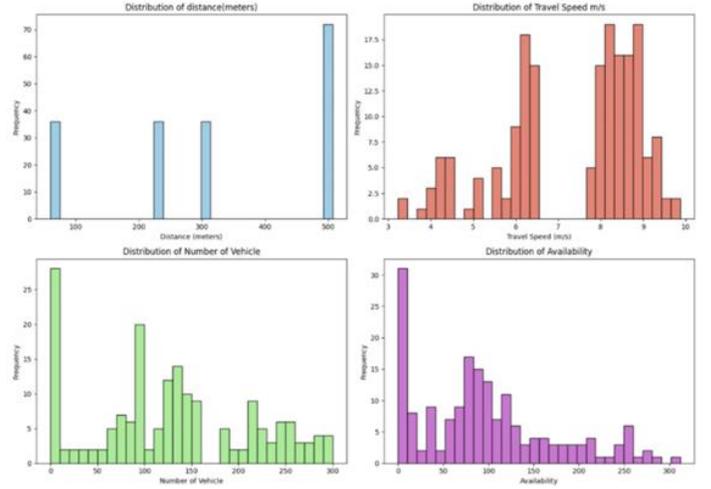

Fig. 6. Frequency of features

The RFR model was configured with 100 decision trees, leveraging its ensemble capabilities to capture complex patterns in the data. The SVR model demonstrated a moderate ability to capture complex patterns but underperformed compared to Linear Regression.

The LSTM model designed for sequential data, consisted of 50 LSTM units, and was trained over 50 epochs with a batch size of 72. The Adam optimizer [16] was employed, known for its adaptive learning rate and efficiency in handling sparse gradients to ensure stable convergence. Performance metrics from model evaluations indicated that RFR outperformed the other models, delivering the most accurate predictions. The results underscore the strength of ensemble methods like RFR in handling non-linear relationships and variability in the dataset, compared to the sequential modeling capabilities of LSTM and the kernel-based approach of SVR.

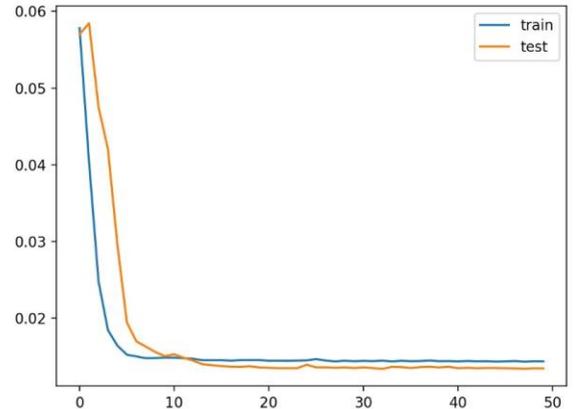

Fig. 7. LSTM Training Graph

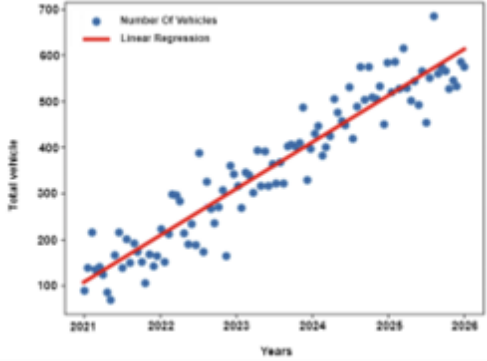

Fig. 8. Linear Regression fitted on number of vehicles

### D. Interactive Parking Assistance Website

To streamline the process of finding available parking spots, we designed and implemented an interactive website that enables students to check real-time parking occupancy across various campus locations. This platform provides a user-friendly interface where students can quickly identify open parking spots as shown in Fig. 9 , Fig. 10 and Fig. 11 allowing them to plan their commute effectively.

The interactive website offers real-time updates on parking availability, enabling students to view the occupancy status of different campus parking areas before they arrive. Through a location-based search feature, users can filter parking spots based on proximity to their classes, meetings, or other destinations, helping them choose the most convenient option. By accessing this information remotely, students can plan their arrival time, accordingly, reducing delays and minimizing unnecessary congestion. The system not only enhances convenience but also encourages more efficient parking management across campus, improving the overall student experience.

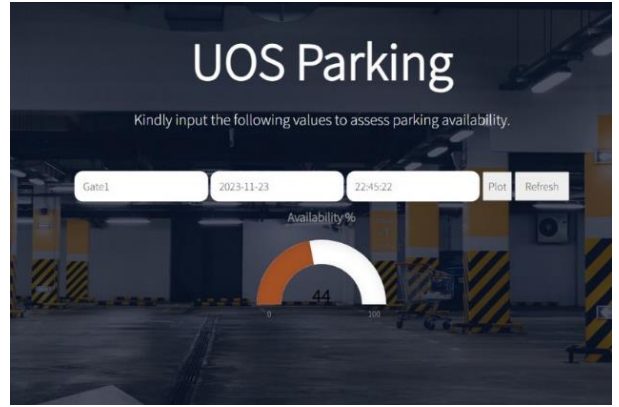

Fig. 10. Moderate occupancy rate

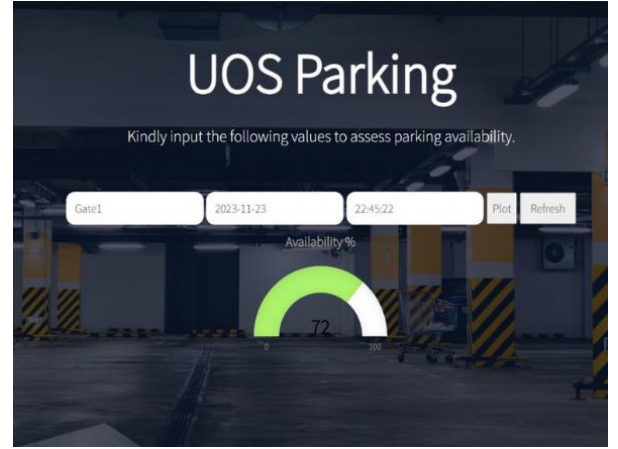

Fig. 11. High occupancy rate

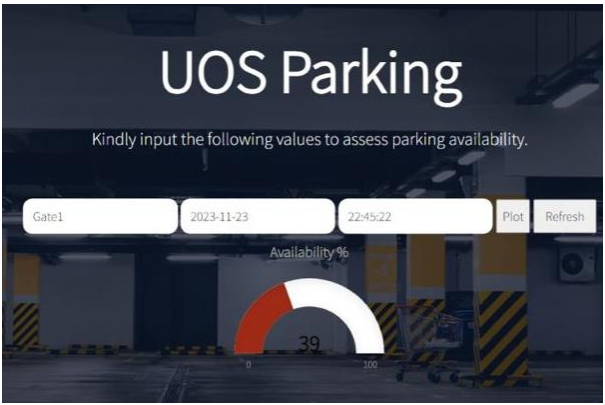

Fig. 9. Low occupancy rate

## IV. RESULTS AND DISCUSSION

The performance of the models was comprehensively evaluated based on three key metrics which are: Mean Absolute Error (MAE) , Root Mean Squared Error (RMSE) and the coefficient of determination ($R^2$). Among the models assessed, the optimal performance was achieved by RFR with the lowest MAE of 0.112, RMSE of 0.142 and highest $R^2$ of 0.582. Corresponding to prediction errors of 10.6 and 13.4 vehicles given the MAE and RMSE obtained.

$$MAE = \frac{1}{n}\sum_{i=1}^{n}|y_i - \hat{y}_i| \qquad (3)$$

The LSTM network achieved a MAE of 0.139, highlighting its potential for temporal sequence modeling tasks, given more data points. In comparison, the SVR model exhibited a RMSE of 0.173 and a MAE of 0.135. While this performance was slightly less effective than the LSTM in terms of RMSE, the SVR model showcased comparable performance in terms of

MAE, suggesting its suitability for regression tasks in specific contexts.

$$RMSE = \sqrt{\frac{1}{n}\sum_{i=1}^{n}(y_i - \hat{y}_i)^2} \quad (4)$$

The Linear Regression model, however, yielded the least favorable results among the tested approaches. It recorded the highest RMSE of 0.208, MAE of 0.178 and lowest $R^2$ of 0.051 indicating limitations in capturing the complexity of the underlying data patterns as shown in Table 1.

TABLE I. COMPARISON OF RESULTS

| Models | Metrics | | |
|---|---|---|---|
| | **RMSE** | **MAE** | **$R^2$** |
| Linear Regression | 0.208 | 0.178 | 0.051 |
| SVR | 0.173 | 0.135 | 0.353 |
| LSTM | 0.149 | 0.139 | 0.457 |
| RFR | 0.142 | 0.112 | 0.582 |

These findings underscore the superior predictive capability of the RFR model and the LSTM network for the dataset, while highlighting areas for improvement in simpler regression techniques such as Linear Regression.

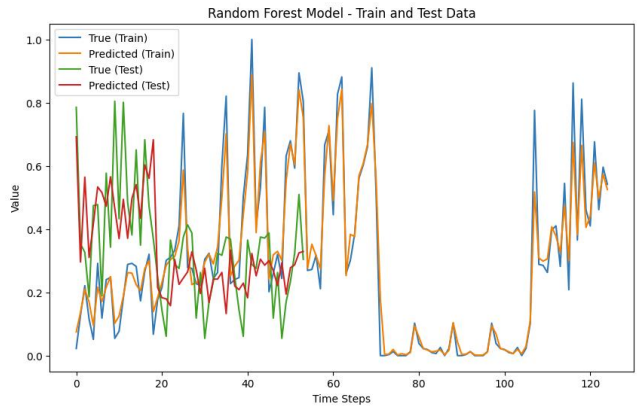

Fig. 12. RFR predicted values compared to actual values

## V. CONCLUSIONS AND FUTURE WORKS

This study demonstrates the potential of integrating predictive analytics and spatial-temporal data to address parking challenges on university campuses. By leveraging geospatial data, vehicle influx and outflux patterns, and machine learning models, the proposed system offers a privacy-preserving and cost-efficient approach to forecasting parking availability. Among the models tested, Random Forest Regression (RFR) emerged as the most effective, achieving the lowest RMSE and MAE values, indicating its robustness in handling complex, non-linear relationships in the dataset. The results underscore the feasibility of utilizing predictive models for real-time parking management, reducing congestion, and optimizing parking allocation. This approach eliminates the need for costly infrastructure such as cameras or sensors, offering a scalable solution for smart campuses and cities. Furthermore, the system's design ensures individual privacy by relying on aggregated vehicular data rather than sensitive personal information. Future work could focus on collecting more data for longer durations and scaling this framework to larger urban environments, integrating real-time data streams for continuous optimization, as well as exploring additional features such as user feedback and dynamic pricing. This research lays the groundwork for developing intelligent transportation systems that enhance user experience, improve traffic management, and contribute to the broader vision of smart city innovation.